\documentclass[11pt]{article}
\usepackage[a4paper,margin=2.5cm]{geometry}
\usepackage[T1]{fontenc}
\usepackage[utf8]{inputenc}
\usepackage{lmodern}
\usepackage{microtype}
\usepackage{amsmath}
\usepackage{booktabs}
\usepackage{graphicx}
\usepackage{xcolor}
\usepackage[hidelinks]{hyperref}
\usepackage{pgfplots}
\pgfplotsset{compat=1.18}
\usepackage[numbers,sort&compress]{natbib}

\title{System Switch: When Should a Fast Decision Model Stop and Think?\\[2pt]
\large A closed-loop study with typed-decision models and a vision reasoner in Doom}
\author{Gianluca Bailo\\ Independent Researcher, Recco (Genoa), Italy\\ \href{mailto:dexmac@libero.it}{\texttt{dexmac@libero.it}}}
\date{October 2026}

\begin{document}
\maketitle

\begin{abstract}
Dual-process agents pair a fast policy with a slow deliberative model. In real-time settings the
slow model usually runs continuously; in turn-based agents and in robot planners it is invoked on
events such as uncertainty or a detected failure. We study a fast learned actor that takes every
decision and hands control to a reasoning vision-language model only when a gate opens, while the
game keeps running in real time. We use closed-loop Doom and the new open ``System One''
typed-decision models, served through a common llama.cpp interface. On 900 held-out questions,
(i) zero-shot decision models from 0.15B to 9B parameters over-choose collecting items, 1.6--1.8 times
more often than chance among their errors, whatever the order of the options, which does change some
models' accuracy;
(ii) accuracy, calibration and sensitivity (how well confidence separates right from wrong answers)
are distinct: models of similar accuracy differ widely in AUROC, and the confidence of the most sensitive
one tracks which kinds of situation it fails rather than which of its answers are wrong;
(iii) offline, deferring the least confident 30\% of decisions to a reasoning model gains over deferring
at random in proportion to the actor's AUROC (rank correlation 0.87 over 18 actor--option-order
pairs); with the actor and the rate chosen on held-out games the gain is +0.13 [0.08, 0.18] with
doomLaya's option order and +0.08 [0.02, 0.14] with shuffled options, and reasoning carries about half
of it.
(iv) In closed loop (33 games on three seeds), no variant reaches the exit. Committing to plans, the
reasoner's or a fixed \emph{explore} rule's behind the same gate, covers more of the map, opens more
doors and even makes an actor that stands still on its own play; with the rule, the agent dies more
often. The thoughts show where the chain breaks: told that some doors need keys, the reasoner takes
ordinary doors for locked ones, which the state cannot tell apart, and without that knowledge it goes
back to collecting. We release code, prompts and logs.
\end{abstract}

\section{Introduction}
The ``System 1 / System 2'' picture \citep{kahneman2011} has become a design template for agents:
a fast reactive component and a slow deliberative one. Turn-based agents switch to the slow
component on events, such as lack of reward, invalid actions or repeated failures
\citep{swiftsage,cogniweb}, choose between fast and slow solvers with a metacognitive module
\citep{sofai}, or route each step to a chosen reasoning depth
\citep{fast,talkerreasoner,cogrouter,pangu}. Robot planners replan when execution feedback or a
detector signals a failure \citep{innermonologue,doremi}. Real-time agents, where the environment does
not wait, keep the slow component running continuously, either asynchronously
\citep{dptagent,agilethinker} or at a fixed rate as in robot policies \citep{groot}.

The starting point of this work is a first-person observation by the author, a long-time player:
when the action works, the brain ``switches off'' and just acts, without inner speech; when
shooting no longer helps (a door that does not open, an enemy that keeps coming back), one stops
to think, and it is the thinking that gets one moving again. This is event-gated thinking
\emph{in real time}: the game goes on while one thinks. It echoes cognitive accounts in which
control is recruited by conflict or lack of progress \citep{botvinick2001} when its expected value
exceeds its cost \citep{shenhav2013}, and the older AI principle of rational metareasoning: compute
only when the computation is worth its cost \citep{russellwefald1991}.

Three developments make the question timely and cheap to study. First, a family of open ``System
One'' typed-decision models appeared in September 2026 (Laya, Kev, Lev, Julia, Clef, OpenJev,
Strands Decider, CLM; \citep{laya,kev,lev,julia1,clef,openjev,strands,clm}). They answer typed
questions (\emph{choice}, \emph{score}, \emph{yes/no}) in a single forward pass, with probabilities
that look calibrated. Second, llama.cpp added a common \texttt{/v1/systemone} endpoint for them
\citep{llamacpp}, so one bridge serves every model on the same hardware. Third, independent
benchmarks of these models \citep{jevindex,jevbench} already locate their weak spot: the items they
fail ``need multi-step inference, arithmetic, or weighing competing considerations'', while ``a
System One model does one forward pass and a linear readout'', and the top of the leaderboard is
held by reasoning models \citep{jevbench}. This suggests a switch rather than a replacement. Those
benchmarks, however, score single decisions. We put the models in a closed loop with doomLaya
\citep{doomlaya}, which turns ViZDoom \citep{vizdoom} into a typed-decision task with a rule-based
expert (an ``oracle'') that labels game states.

\paragraph{Contributions.}
\begin{enumerate}
\item \textbf{A fast learned actor that asks a reasoner only when a gate opens, in real time.} Prior
real-time agents keep the slow component running continuously \citep{dptagent,agilethinker}; most
event-gated agents are turn-based \citep{swiftsage,cogniweb,sofai}; robot planners that replan on
detected failures \citep{innermonologue,doremi} take every high-level decision with the language
model. Here a fast typed-decision actor takes every decision, and a vision reasoner is consulted,
asynchronously, only when uncertainty or progress triggers open a gate; we compare the two kinds of
trigger in closed loop (\S\ref{sec:switch}, \S\ref{sec:loop}).
\item \textbf{What the confidence of typed actors tracks, and what deferral buys.} We evaluate the
System One family zero-shot through one interface and show that accuracy, calibration and
sensitivity (the AUROC of confidence for right vs.\ wrong answers) are distinct properties, that the
confidence of the most sensitive actor tracks \emph{which kinds} of situation it fails rather than which answers, and
what confidence-gated deferral to a reasoner buys offline, with the actor and the deferral rate
chosen on held-out games (\S\ref{sec:zeroshot}--\S\ref{sec:defer}).
\item \textbf{A closed-loop study with a rule control and a reading of the thoughts.} The same gate
with a fixed command instead of thinking, ablations of the image and of the game knowledge given to the
reasoner, and a reading of every verbalised thought (\S\ref{sec:loop}).
\end{enumerate}

\section{Related work}
\paragraph{Fast and slow agents.}
SwiftSage \citep{swiftsage} pairs a small imitation-learned actor with an LLM planner and switches
on heuristic triggers, including lack of reward and invalid actions; CogniWeb \citep{cogniweb} uses
similar stuck and invalid-state triggers for web agents. FAST \citep{fast}, Talker--Reasoner
\citep{talkerreasoner}, CogRouter \citep{cogrouter} and Pangu Embedded \citep{pangu} select or adapt
reasoning depth per step. SOFAI \citep{sofai} has a metacognitive module that chooses, problem by
problem, between a fast solver and a slow one using the fast solver's confidence and the expected cost
of thinking, which is the offline question of our \S\ref{sec:defer}. All of these operate turn by
turn. In robotics, Inner Monologue \citep{innermonologue} feeds execution feedback (success detection,
scene descriptions) back to a language planner, and DoReMi \citep{doremi} has a vision-language model
watch the execution for violated constraints and triggers replanning as soon as one is detected: event
gating in a continuous environment, with the language model as the planner of every step. Real-time settings have been addressed
by DPT-Agent \citep{dptagent}, which runs an asynchronous reflective System~2 alongside a finite-state
System~1 in simultaneous human--AI collaboration, and by AgileThinker \citep{agilethinker}, which runs a reactive and a planning thread in parallel,
the reactive one reading the partial reasoning traces of the other, in environments that evolve
regardless of the agent's computation. In both, the slow component runs continuously. Dual-system robot policies such
as GR00T~N1 \citep{groot} run it at a fixed rate. Our setting differs from all of these in where the
decisions come from: a fast learned actor takes every decision, the reasoner is invoked only when a
gate opens, and the game keeps running while it thinks.
\paragraph{Deferral, selective prediction and cascades.}
Abstaining when uncertain is the subject of selective classification \citep{geifman2017}; learning
when to pass a decision to another decision maker is learning to defer \citep{madras2018}; routing
queries through models of increasing cost is the idea of LLM cascades \citep{frugalgpt}. Closest in
the agent domain, ReDAct \citep{redact} lets a small LLM act and defer to a large one when its
uncertainty exceeds a calibrated threshold; deferring about 15\% of the decisions matches the large
model on ALFWorld and MiniGrid. We differ in that our actors are non-generative typed-decision models
whose confidence must first be shown to be informative, and in that we gate on progress as well as
on uncertainty, in real time.
\paragraph{System One models and their benchmarks.}
The Jev Decision Index \citep{jevindex} compares open reproductions of the TypeSafe Jev model on
dozens of decision tasks; JevBench \citep{jevbench} is an independent benchmark on which reasoning
models lead. Both score single decisions; we add a closed loop.
\paragraph{Doom.}
ViZDoom \citep{vizdoom} is the standard platform; recent work shows that a 1.3M-parameter specialized
model outplays general LLMs in real-time control \citep{doom13m}. We use doomLaya \citep{doomlaya},
which exposes Doom as typed questions over a symbolic state, with an executor that turns the chosen
command into movement.

\section{Setup}\label{sec:setup}
\paragraph{Environment and questions.}
Freedoom~2 \citep{freedoom} in ViZDoom 1.3.1 through doomLaya (commit \texttt{b25edd3}). Every 0.5\,s
of game time the agent receives a text state (health, armour, inventory, visible enemies, items,
doors) and two \emph{choice} questions: \emph{command} (attack a visible enemy, collect an item, open
a door, go to the exit, retreat, explore, wait) and \emph{weapon}. The options come in a fixed order
(enemies, items, door, exit, retreat, explore, wait). An executor turns the command into button presses
and reports whether it is executing, arrived, blocked or unavailable. The state reports a closed door
but not whether it is locked, and a key only once it has been seen within 12\,m.
\paragraph{Clock.}
ViZDoom runs in synchronous mode, paced by doomLaya to real time: 35 tics per second, and the
measured wall-clock duration of every game equals its game time. A decision is requested every
0.5\,s of game time, asynchronously: while an answer is pending, the game goes on with the previous
command. The switch answers the actor at once and thinks in a background thread, so a thought that
takes 12\,s takes 12\,s of play.
\paragraph{Labels.}
doomLaya's oracle (the rule-based labeller) labels states with rules (fight visible threats, collect
only what is needed, open blocking doors, otherwise exit). It never labels \emph{explore},
\emph{retreat} or \emph{wait}: they are offered but never correct, so the offline metric cannot reward
exploring, which is what the thinker often prescribes in the loop. Offline accuracy and closed-loop
behaviour therefore measure different things. Its validation split has 130 questions (80 command, 50 weapon). We
build a held-out test set of 900 questions with the same pipeline and the same oracle on 42 MAP01
games played with seeds never used for training (45--53), capped at 150 per (question, category)
group (\texttt{build\_testset.py}). Confidence intervals on this set resample games, not questions,
because states from the same game are correlated. ``Accuracy'' means agreement with the oracle; we
come back to what this hides in \S\ref{sec:limits}.
\paragraph{Actors.}
Zero-shot System One models through llama.cpp's \texttt{/v1/systemone} (commit \texttt{cb7934c}):
Julia-1 (0.15B), Laya (0.4B), Kev (0.8B, 4B, 9B), Lev (4B), Clef-Flash (9B); Strands Decider 2B through
its own server with the same API. Laya is evaluated in two released revisions, the
typed-decisions model and the English base, through doomLaya's PyTorch server (the base also
through llama.cpp). Task-trained references: Laya-v3 (doomLaya's fine-tune) and
doomGemma, a Gemma-4-E2B LoRA trained on the same 979 examples \citep{doomgemma}.
\paragraph{Thinkers.}
Two vision LLMs with official weights, served by llama.cpp with their vision projectors, each
with and without reasoning: Qwen3.6-35B-A3B (4-bit GGUF with its multi-token-prediction head) and
Gemma~4 26B-A4B-it (quantization-aware 4-bit). Offline, the thinkers see the same text state as the
actors, without the frame; in the closed loop, which uses Qwen3.6, the thinker also receives the
current frame. Answers are
constrained to the option keys by a JSON schema; reasoning is capped at 4{,}000 tokens, and an answer
that does not arrive within the cap counts as wrong.
\paragraph{Hardware.}
Home PCs on gigabit Ethernet: the game and the actors on an RTX 4070 / RTX 3060 machine, Qwen3.6 on 2$\times$RTX 5060 Ti, Gemma~4 on the RTX 4070 / RTX 3060; System One baselines
on a Ryzen 9 CPU or 2$\times$RTX 2070 SUPER (accuracy does not depend on the device); the rule control
with the base Laya was played on the 2$\times$RTX 2070 SUPER machine.

\section{The System Switch}\label{sec:switch}
\texttt{switch.py} sits between doomLaya's agent and the actor, with the same interface. The actor
answers every request. A \emph{gate} decides whether to think, from five signals on the recent
history: \emph{unsure} (top probability below a threshold for three decisions in a row, the
ReDAct-style trigger; 0.3 for the typed-decisions Laya and 0.65 for the base, the same percentile of
their confidence on the test set),
\emph{stuck} (the player moved less than 48 map units in 5\,s with no enemy in sight), \emph{blocked}
(three of the last six commands could not be executed), \emph{loop} (the same command in $\geq$80\% of
the decisions of the last 15\,s), \emph{novelty} (no new 128-unit cell reached in 15\,s). The last four
are progress triggers in the sense of \citep{swiftsage,cogniweb}. When the gate opens, the thinker
receives the game frame, the state, the available commands, the recent history and its own previous
thoughts, writes one first-person thought and picks a command, which then overrides the actor (a
\emph{plan}). The game keeps running while it thinks. Variants: \emph{rule} (same gate, a fixed
command instead of thinking: the control), \emph{advise} (the thought is appended to the actor's state
instead of overriding it), \emph{basics} (a few sentences of Doom knowledge in the thinker's prompt),
\emph{forbid-repeats} (the commands the actor kept repeating are removed from the thinker's options),
\emph{commit} (a plan lasts until the executor reports it done or impossible, a damage burst, or 45\,s).

\section{Results}
\subsection{Zero-shot decision models over-choose collecting}\label{sec:zeroshot}
\begin{table}[h]\centering\footnotesize\setlength{\tabcolsep}{4pt}
\begin{tabular}{lrcccc}\toprule
& & \multicolumn{2}{c}{command} & & \\
model & params & fixed order & shuffled & weapon & $\rightarrow$collect ($\times$chance)\\\midrule
Kev & 9B & 0.60 [0.49, 0.69] & 0.50 & 1.00 & 92\% (1.7)\\
Strands Decider & 2B & 0.59 [0.49, 0.68] & 0.38 & 1.00 & 97\% (1.8)\\
Laya, typed-decisions & 0.4B & 0.59 [0.48, 0.68] & 0.56 & 0.42 & 91\% (1.6)\\
Laya, base & 0.4B & 0.56 [0.45, 0.66] & 0.49 & 0.38 & 83\% (1.7)\\
Clef-Flash & 9B & 0.54 [0.43, 0.63] & 0.54 & 0.60 & 92\% (1.7)\\
Lev & 4B & 0.43 [0.31, 0.55] & 0.44 & 0.49 & 98\% (1.8)\\
Kev & 4B & 0.40 [0.31, 0.52] & 0.39 & 0.97 & 98\% (1.8)\\
Kev & 0.8B & 0.29 [0.22, 0.39] & 0.39 & 0.05 & 56\% (1.1)\\
Julia-1 & 0.15B & 0.14 [0.07, 0.24] & 0.09 & 0.74 & 91\% (1.8)\\
\midrule
Qwen3.6-35B-A3B, no reasoning & 35B (3B act.) & 0.58 [0.48, 0.67] & -- & -- & 97\% (1.8)\\
Qwen3.6-35B-A3B, reasoning & 35B (3B act.) & 0.69 [0.60, 0.76] & -- & -- & 73\% (1.4)$^\dagger$\\
Gemma~4 26B-A4B, no reasoning & 26B (4B act.) & 0.57 [0.47, 0.65] & -- & -- & 90\% (1.7)\\
Gemma~4 26B-A4B, reasoning & 26B (4B act.) & 0.50 [0.39, 0.60] & -- & -- & 94\% (1.6)$^\dagger$\\
\midrule
Laya-v3 (trained) & 0.4B & 0.59 [0.47, 0.70] & 0.59 & 0.97 & 61\% (1.2)\\
doomGemma (trained) & 2B & 0.86 [0.81, 0.90] & 0.85 & 1.00 & 43\% (0.8)\\\bottomrule
\end{tabular}
\caption{Accuracy against the oracle on the 900 held-out questions (600 command, balanced over
attack, collect, open door and exit; 300 weapon), with 95\% intervals that resample games.
\emph{Fixed order}: doomLaya's order of the options, the one the agent sees in play; \emph{shuffled}:
options in a random order per question. $\rightarrow$collect: share of the command errors that choose an
item to collect, and in brackets its ratio to the share expected if the wrong answer were uniform over
the wrong options. The thinkers were asked the command questions in the fixed order only; with
reasoning, an answer not given within 4{,}000 tokens counts as wrong (Qwen3.6: 13 of 600; Gemma~4:
199). $^\dagger$Among the errors with an answer.}
\label{tab:zeroshot}
\end{table}
Every zero-shot model but one over-chooses collecting (Table~\ref{tab:zeroshot}). On average a command
question has 7.5 options, 3.6 of them items to collect, so many errors would land on an item by chance;
yet the zero-shot models choose one 1.6--1.8 times more often than a uniformly random wrong answer would.
The exception is Kev 0.8B (1.1), whose probabilities are near-uniform. The thinkers share the bias
without reasoning (1.8) and less with it (1.4 for Qwen3.6). By category, every zero-shot model rarely
leaves when the oracle leaves (0.00--0.35) and rarely picks the item the oracle picks (0.15--0.55); the
models differ on fighting (0.13--0.98) and on doors.

The order of the options matters for some models. doomLaya lists them in a fixed order, and the
correct answer is the first option in 41\% of the command questions. With the options shuffled per
question, Strands drops from 0.59 to 0.38 (it chooses the first option in 65\% of the questions in the
fixed order and still in 36\% when shuffled, where the correct one is first in 14\%), Kev 9B from 0.60
to 0.50 and the Laya base from 0.56 to 0.49, while Clef-Flash is unaffected and Kev 0.8B improves. On
weapons, Kev 4B drops from 0.97 to 0.62. The over-choice of collecting is not an effect of position:
with shuffled options it stays at 1.6--1.8 times chance for all the models above, and only Julia-1's
disappears (0.9). Kev 0.8B answers \emph{melee} to 95\% of the weapon questions whatever their position,
which explains its weapon score. The thinkers saw the fixed order only.

On the balanced set, accuracy grows with size within the Kev family (0.29, 0.40, 0.60). On the
130-question validation split, which has only 5 door states among its 80 command questions, the order was
reversed: Kev 0.8B and 4B never answer \emph{open door} correctly (0 of 150 each). Task training transfers
less than the validation split suggests. doomGemma drops from 0.963 on the validation split, on which its
epoch was selected, to 0.860 on questions from unseen seeds, consistent with its failure to generalise to
an unseen map \citep{doomgemma}. doomLaya's own fine-tune, Laya-v3, drops from 0.725 to 0.585, no better
than zero-shot Laya, mainly because it collects items at doors (0.39 on \emph{open door}). The trained
models are insensitive to the order of the options (0.85 and 0.59 shuffled). The llama.cpp conversion is
faithful: the Laya base through \texttt{/v1/systemone} scores as the PyTorch original, on the 130
questions (0.41/0.58) and on the 900 (0.562 on commands, AUROC 0.888 vs.\ 0.891; shuffled, 0.495 vs.\
0.492).

\subsection{Calibration, sensitivity, and what the confidence tracks}\label{sec:meta}
System One models ship temperature-scaled probabilities: the temperatures are stored in the model
file and applied by the server, which warns that the result is not guaranteed to be calibrated for
one's data \citep{llamacpp}; Strands Decider stores one temperature per question type \citep{strands}.
A confidence-gated switch needs something else than calibration: that confidence \emph{separates}
right from wrong answers, i.e.\ metacognitive sensitivity \citep{fleming2014}. We measure calibration
as the expected calibration error (ECE, 15 bins) of the top probability, and sensitivity as the AUROC
of the top probability for right vs.\ wrong answers, overall and within each gold category (only
right/wrong pairs of the same category count).
\begin{table}[h]\centering\small
\begin{tabular}{lccccc}\toprule
& & & \multicolumn{2}{c}{AUROC, fixed order} & AUROC\\
actor & accuracy & ECE & overall & within category & shuffled\\\midrule
Laya, base & 0.56 & 0.16 & 0.89 [0.85, 0.92] & 0.52 [0.40, 0.63] & 0.72\\
Kev 0.8B & 0.29 & 0.14 & 0.83 [0.76, 0.88] & 0.75 [0.63, 0.85] & 0.77\\
Laya, typed-decisions & 0.59 & 0.29 & 0.77 [0.70, 0.83] & 0.65 [0.42, 0.81] & 0.69\\
Kev 4B & 0.40 & 0.10 & 0.70 [0.58, 0.80] & 0.80 [0.68, 0.89] & 0.66\\
Clef-Flash 9B & 0.54 & 0.10 & 0.68 [0.59, 0.76] & 0.68 [0.58, 0.78] & 0.68\\
Lev 4B & 0.43 & 0.24 & 0.61 [0.51, 0.70] & 0.61 [0.46, 0.75] & 0.63\\
Kev 9B & 0.60 & 0.28 & 0.61 [0.48, 0.71] & 0.74 [0.64, 0.81] & 0.70\\
Strands Decider 2B & 0.59 & 0.21 & 0.60 [0.47, 0.71] & 0.77 [0.67, 0.85] & 0.52\\
Julia-1 & 0.14 & 0.57 & 0.45 [0.34, 0.57] & 0.58 [0.39, 0.75] & 0.44\\
\midrule
Laya-v3 (trained) & 0.59 & 0.14 & 0.87 [0.83, 0.91] & 0.78 [0.57, 0.89] & 0.79\\
doomGemma (trained) & 0.86 & 0.05 & 0.86 [0.80, 0.90] & 0.77 [0.68, 0.84] & 0.84\\\bottomrule
\end{tabular}
\caption{Calibration and sensitivity on the 600 command questions of the held-out set: ECE of the top
probability, and AUROC of the top probability for right vs.\ wrong answers (0.5 = chance), overall and
within gold categories, with 95\% intervals that resample games; last column: overall AUROC with the
options shuffled. Zero-shot models above the line, sorted by AUROC.}\label{tab:meta}
\end{table}
Accuracy, calibration and sensitivity are distinct properties (Table~\ref{tab:meta}). Five zero-shot
models lie within 0.54--0.60 command accuracy, and their AUROC ranges from 0.60 to 0.89; the two most
accurate, Kev 9B and Strands, have the least informative confidence of the five, with intervals that
include chance. The two Laya revisions, equally accurate, differ (0.77 vs.\ 0.89, non-overlapping
intervals). Calibration does not predict sensitivity: the best-calibrated zero-shot models, Clef-Flash
and Kev 4B (ECE 0.10), rank in the middle by AUROC, and the typed-decisions Laya, Strands and Kev 9B are
under-confident (mean top probability 0.31--0.42 for accuracies of 0.59--0.60).

Sensitivity itself comes in two kinds. The base Laya, the most sensitive overall, is at chance within
categories (0.52 [0.40, 0.63]): its confidence is high on fights and doors, which it gets right, and
low on exits and collects, which it gets wrong, but within a category it does not track which of its
answers are wrong. Strands and Kev 9B show the opposite profile (0.77 and 0.74 within categories, 0.60
and 0.61 overall). What pays in the deferral of \S\ref{sec:defer} is the first kind, because accuracy
differs so much between categories (from 0.00 on exits to 0.98 on fights for the base Laya): a
confidence that tracks which situations the actor handles badly, rather than which answers are wrong. ``Knowing when one does not
know'' is therefore more than the overall AUROC shows. An actor for a confidence-based switch should be
chosen by the AUROC that matters for the deferral, not by accuracy. Type-2 AUROC also depends on
type-1 performance \citep{fleming2014}: comparisons among the five models of similar accuracy are fair,
comparisons across the Kev family (0.29 to 0.60) are not, and we draw none. These estimates replace those
of our 130-question pilot, which were too noisy to rank models: there the two Laya revisions were ordered
the other way (0.87 vs.\ 0.82, overlapping intervals) and Strands looked worse than chance (0.41 [0.29,
0.54]), with an interval that includes it.

Part of the sensitivity measured with the fixed order comes from the order itself. With shuffled
options the spread narrows (0.52--0.77 among the zero-shot models): the Laya base falls to 0.72, Kev 9B
rises to 0.70, and Strands stays at chance (0.52). The dissociation from accuracy holds (Clef-Flash,
the most accurate with shuffled options, ranks in the middle), but ``the two most accurate are the least
sensitive'' is a property of the fixed order.

\subsection{Offline: confidence-gated deferral to a reasoner}\label{sec:defer}
We defer the actor's least confident decisions to the thinker and measure the accuracy of the
combination for each deferral rate, against deferring at random. This is the selective-prediction /
cascade setting \citep{geifman2017,madras2018,frugalgpt,redact}, measured offline on the labelled
questions, i.e.\ with all the time the thinker needs.
\begin{table}[h]\centering\footnotesize\setlength{\tabcolsep}{4pt}
\begin{tabular}{lccccc}\toprule
& & \multicolumn{3}{c}{deferred to the thinker} & confidence $-$ random\\
actor (AUROC) & alone & 10\% & 30\% & 50\% & at 30\%\\\midrule
\multicolumn{6}{l}{\emph{thinker: Qwen3.6 with reasoning (alone: 0.69 [0.60, 0.76])}}\\
Laya, base (0.89) & 0.56 & 0.61 & 0.75 & 0.76 & +0.15 [0.11, 0.19]\\
Laya-v3, trained (0.87) & 0.59 & 0.62 & 0.72 & 0.82 & +0.11 [0.06, 0.18]\\
Laya, typed-decisions (0.77) & 0.59 & 0.60 & 0.67 & 0.72 & +0.05 [0.02, 0.08]\\
Kev 9B (0.61) & 0.60 & 0.60 & 0.62 & 0.63 & $-$0.01 [$-$0.04, 0.01]\\
Strands Decider (0.60) & 0.59 & 0.58 & 0.60 & 0.63 & $-$0.02 [$-$0.06, 0.02]\\
doomGemma, trained (0.86) & 0.86 & 0.86 & 0.84 & 0.83 & +0.03 [0.01, 0.06]\\
\multicolumn{6}{l}{\emph{thinker: Qwen3.6 without reasoning (alone: 0.58 [0.48, 0.67])}}\\
Laya, base (0.89) & 0.56 & 0.57 & 0.64 & 0.63 & +0.07 [0.03, 0.12]\\
\multicolumn{6}{l}{\emph{thinker: Gemma~4 with reasoning (alone: 0.50 [0.39, 0.60])}}\\
Laya, base (0.89) & 0.56 & 0.57 & 0.67 & 0.65 & +0.13 [0.08, 0.17]\\\bottomrule
\end{tabular}
\caption{Offline deferral on the 600 command questions of the held-out set, in doomLaya's option order: the actor's least
confident decisions are answered by the thinker. Last column: accuracy gained by deferring by
confidence rather than at random, with a 95\% interval that resamples games. For reference, a
perfect selector that defers exactly the Laya base's mistakes reaches 0.76 at 30\%.}\label{tab:defer}
\end{table}
\begin{figure}[h]\centering
\begin{tikzpicture}
\begin{axis}[width=0.82\linewidth, height=6.2cm, xmin=0, xmax=100, ymin=0.5, ymax=0.9,
  xlabel={decisions deferred to the thinker (\%)}, ylabel={command accuracy},
  legend style={font=\scriptsize, cells={anchor=west}, at={(0.5,-0.22)}, anchor=north, legend columns=2},
  tick label style={font=\footnotesize}, label style={font=\small}, grid=major, grid style={gray!20}]
\addplot[very thick, color={rgb,255:red,0;green,90;blue,181}] table[x=pct, y=conf]{fig/defer_laya-base.dat};
\addlegendentry{Laya base (AUROC 0.89), by confidence}
\addplot[thick, dashed, color={rgb,255:red,0;green,90;blue,181}] table[x=pct, y=random]{fig/defer_laya-base.dat};
\addlegendentry{Laya base, at random}
\addplot[thick, dotted, color={rgb,255:red,0;green,90;blue,181}] table[x=pct, y=perfect]{fig/defer_laya-base.dat};
\addlegendentry{Laya base, perfect selector}
\addplot[thick, color={rgb,255:red,86;green,180;blue,233}] table[x=pct, y=conf]{fig/defer_laya-typed.dat};
\addlegendentry{Laya typed-decisions (0.77)}
\addplot[thick, color={rgb,255:red,230;green,159;blue,0}] table[x=pct, y=conf]{fig/defer_kev-9b.dat};
\addlegendentry{Kev 9B (0.61)}
\addplot[thick, dashdotted, color={rgb,255:red,204;green,121;blue,167}] table[x=pct, y=conf]{fig/defer_strands-2b.dat};
\addlegendentry{Strands Decider (0.60)}
\addplot[thin, gray, forget plot] coordinates {(0,0.86) (100,0.86)} node[pos=0.98, font=\scriptsize, anchor=north east] {doomGemma, trained: 0.86};
\end{axis}
\end{tikzpicture}
\caption{Accuracy of actor + Qwen3.6 with reasoning on the 600 command questions (doomLaya's option order), for every deferral
rate (the thinker alone is the right end, 0.69). Deferring by confidence beats deferring at random only
for the actors whose confidence is informative.}\label{fig:defer}
\end{figure}
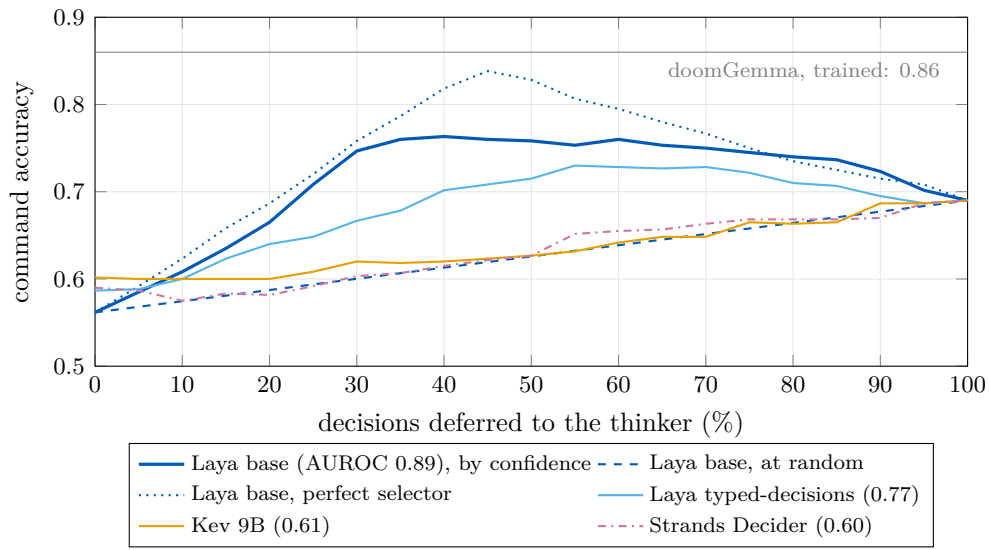
Deferral pays where the actor's confidence is informative (Table~\ref{tab:defer},
Figure~\ref{fig:defer}). With the reasoning thinker, deferring the 30\% least confident decisions of
the Laya base raises command accuracy from 0.56 to 0.75 [0.63, 0.82], close to a perfect selector that
defers exactly the actor's mistakes (0.76), without any training. This matches the thinker alone
(0.69; difference +0.06 [$-$0.02, 0.11]) while calling it on 30\% of the decisions. The gain over random
deferral follows the actors' AUROC: +0.15 for the Laya base (0.89), +0.11 for Laya-v3 (0.87), +0.05 for
the typed-decisions Laya (0.77), and none for Kev 9B and Strands (0.61, 0.60). With those two,
deferring by confidence is deferring at random, although they are the most accurate zero-shot actors.
Laya-v3, trained on the task but no better than zero-shot Laya on this set, reaches 0.82 when half of
its decisions are deferred, close to doomGemma (0.86): its confidence is informative (AUROC 0.87) even
though its answers are not better.

The Laya base and the 30\% rate are chosen here on the same 600 questions on which the gain is
measured, out of nine candidate actors (the zero-shot actors of Table~\ref{tab:meta}; the trained ones
are excluded) and several rates. To check how much this flatters the result,
we split the 42 games into two halves 500 times at random, choose the actor (by AUROC) and the rate
(10--50\%) on one half and measure on the other. The Laya base is chosen in 489 of the 500 splits; on
the held-out half the gain over random deferral is +0.13 [0.08, 0.18], while the advantage over the
better of actor and thinker alone is +0.05 [$-$0.12, 0.11]. The robust result is therefore the gain over
random deferral, and the thinker's accuracy at a fraction of its calls, not a combination that beats
both components.

The size of the gain also depends on the order of the options. With the actors' answers to shuffled
options (the thinker's answers are those to the fixed order), the Laya base gains +0.07 [0.04, 0.09]
over random at 30\% instead of +0.15, and Strands still nothing. Choosing actor and rate on held-out
games as above (the choice now falls mostly on Kev 0.8B), the gain over random is +0.08 [0.02, 0.14], and
the combination stays below the thinker alone ($-$0.05 [$-$0.13, 0.02]). What holds in both orders is the
link with sensitivity: over the same nine zero-shot actors in the two orders (18 actor--order pairs,
not independent), the rank correlation between an actor's AUROC and its gain over random deferral at
30\% is 0.87.

Reasoning carries about half of the gain: without it the thinker (0.58) shares the actors' bias, with
97\% of its errors choosing an item to collect, and the gain over random falls from +0.15 to +0.07. The
result holds with the second thinker: Gemma~4 is weaker alone (0.50; it gives no answer within the
reasoning budget on a third of the questions), yet deferring 30\% of the Laya base's decisions to it
gives 0.67, +0.13 [0.08, 0.17] over random. For the trained doomGemma, deferral to a weaker thinker never
raises accuracy above the actor alone, although it beats deferring at random. These are offline
numbers: the reasoning thinker writes a median of 1{,}300 tokens per decision, and with one decision
every 0.5\,s deferring 30\% of them is not possible in real time. This is what motivates gating on
events and letting a thought guide several decisions (\S\ref{sec:loop}).

\subsection{Closed loop: a cascade of bottlenecks (case study)}\label{sec:loop}
We played MAP01 on seeds 48--50 with Qwen3.6 with reasoning as thinker and two actors on a GPU (a
median of 29\,ms per decision, 350 decisions per 180-s game). The main actor is the typed-decisions
Laya, chosen on the 130-question pilot, where it ranked first by AUROC; on the held-out set the base
ranks first (\S\ref{sec:meta}), so we add it as a second actor. The variants are: \emph{off}; a
\emph{bare} thinker (all five triggers, no basics, no commitment); three variants with basics,
forbid-repeats, plans of at least 5\,s and commitment, which differ only in the gate (uncertainty only,
progress only, or all five triggers); the last of these without the frame (\emph{no image}) and
without the basics (\emph{no basics}); and the \emph{rule} control, with the same gate and the same
commitment but a fixed \emph{explore} instead of a thought. Thirty games in all, plus the rule control
with the base Laya, played later on the 2$\times$RTX 2070 SUPER machine. Since no game reached
the exit, Table~\ref{tab:loop} also reports how much of the map was covered (distinct 128-unit cells,
the grid of the \emph{novelty} trigger), the doors opened and the closest straight-line approach to
the exit.
\begin{table}[h]\centering\footnotesize\setlength{\tabcolsep}{3.5pt}
\begin{tabular}{lcccccccl}\toprule
variant & kills & deaths & cells & doors & closest & thoughts & guided & thinker's top choice\\\midrule
\multicolumn{9}{l}{\emph{actor: Laya, typed-decisions}}\\
off & 11.0 & 0.7 & 77 & 5.3 & 12.6\,m & -- & -- & --\\
bare thinker & 9.3 & 0.7 & 61 & 5.0 & 12.8\,m & 14.3 & 7\% & collect (49\%)\\
commitment, gate: unsure & 12.0 & 2.0 & 66 & 8.0 & 12.9\,m & 7.0 & 28\% & explore (52\%)\\
commitment, gate: progress & 14.3 & 1.7 & 83 & 8.7 & 12.6\,m & 4.0 & 18\% & explore (67\%)\\
commitment, gate: all & 14.3 & 1.3 & 83 & 8.0 & 9.5\,m & 7.0 & 26\% & explore (57\%)\\
\quad no image & 9.7 & 1.3 & 64 & 8.7 & 12.7\,m & 5.3 & 18\% & explore (56\%)\\
\quad no basics & 15.3 & 1.7 & 69 & 8.3 & 10.8\,m & 9.7 & 11\% & collect (45\%)\\
rule control (fixed explore) & 10.7 & 2.7 & 83 & 9.7 & 12.9\,m & (14.0) & 77\% & (explore)\\
\multicolumn{9}{l}{\emph{actor: Laya, base}}\\
off & 0.0 & 0.0 & 1 & 0.0 & 18.8\,m & -- & -- & --\\
commitment, gate: all & 10.7 & 0.7 & 66 & 8.7 & 13.0\,m & 9.3 & 32\% & explore (75\%)\\
rule control (fixed explore)$^*$ & 9.7 & 3.7 & 81 & 9.0 & 13.0\,m & (15.7) & 79\% & (explore)\\\bottomrule
\end{tabular}
\caption{Closed loop on MAP01, three seeds per variant, means per 180-s game. No game reached the exit.
Cells: distinct 128-unit cells visited; closest: closest straight-line distance to the exit (the start is
18.8\,m away); thoughts: thoughts per game (rule: rule activations); guided: share of decisions taken
by a plan. $^*$Played on the 2$\times$RTX 2070 SUPER machine (24\,ms per decision); a check game of the base Laya
alone there was identical to the one on the first machine.}\label{tab:loop}
\end{table}

No variant reaches the exit (Table~\ref{tab:loop}); the route passes through ordinary doors, and only one
game (all triggers, seed 49) came within 4\,m of it. Committing to the thinker's plans covers more of
the map than the bare thinker (64--83 cells against 61) and opens more doors (8--9 against 5). The
\emph{rule} control does as well on both (83 cells, 9.7 doors) and dies twice as often (2.7 deaths per
game against 1.3). With three games per variant each difference in deaths could be noise, but it goes
the same way with the base Laya (3.7 against 0.7). On coverage and
doors, then, most of the benefit of the committed variants comes from committing to explore, not from
what the thinker says. With the typed-decisions actor, deaths also rise from the uncommitted variants (0.7)
to the committed ones (1.3--2.0): a plan keeps the agent moving into areas where enemies are; we did not test this further.
The gate changes \emph{when} the thinker is called (the uncertainty gate opens mostly on low confidence,
the progress gate on failed plans), not these outcomes. The base Laya, which on its own stays still for
the whole game in all three seeds (one cell), plays when the gate is added: 66 cells, 8.7 doors, 10.7
kills, with the gate opened mostly by its low confidence (17 of 28 thoughts). It plays with the rule
control too, and covers even more ground (81 cells, 9 doors): what unblocks it is committing to explore,
not the thinker, whose contribution here is to survival (0.7 deaths per game against 3.7).

The thoughts show where the chain breaks. The bare thinker prescribes collecting the nearest item in
half of its thoughts, and its plans last about a second. With the basics it prescribes exploring, but
its plans then fail at doors: all 25 plan failures of the three gate variants were a closed door
stopping the executor, whose report explicitly suggested \emph{open\_door}. The thinker chose to
open it once; in 17 of the 25 thoughts it concluded that the door needed a key it did not have
(\emph{``Blocked by a closed door. No key. I need to find one. I'll explore the unexplored room to the
left.''}). Without the basics, none of the thoughts after a door failure mentions a key, but the thinker
goes back to collecting (45\% of its thoughts). The basics, which we had added to fix MAP02, where a door
does need the yellow key, trade one wrong prescription for another, and the state cannot settle it: it
reports a closed door, not whether it is locked. Without the frame the thinker reasons longer (a median
of 29\,s per thought; 9 of 16 thoughts exceeded the 4{,}000-token budget and were answered without
reasoning) and covers less (64 cells).

Thinking is expensive in game time. With basics and commitment the thinker is busy for 38--70\% of the
game, and a plan starts on a frame that is a median of 9--29\,s old by the time it applies (4\,s for the
bare thinker, 6\,s without the basics).

A single-seed pilot on MAP01 (Laya) and MAP02 (doomGemma), with the abliterated thinker, had shown the
same progression; at each of its steps we changed several settings at once, so it cannot attribute an
effect to a single one:
\begin{itemize}
\item \textbf{Bare thinker} (5--6\% of decisions guided): the diagnosis is often right
(\emph{``I'm stuck in a shooting loop''}, \emph{``the RedCard is unreachable''}, \emph{``I've been trying
to open the door for 5 seconds, but it needs a key I don't have''}) but the prescription is not: 22 of
24 thoughts end in ``collect the nearest item to break the loop''; \emph{explore} is never chosen.
\item \textbf{+ basics, forbid-repeats, plans $\geq$5\,s} (9--25\%): the prescription becomes right on
MAP02 (\emph{``the door needs a yellow key, which I don't have; I need to explore to find it''}, 12 of 13
thoughts).
\item \textbf{+ commitment} (12--32\%): plans last 26--30\,s and chain (\emph{exit} $\rightarrow$
blocked by a door $\rightarrow$ \emph{open\_door}); they end on damage bursts, when reflexes take over.
\item \textbf{Advise} instead of override: the trained actor ignores the advice.
\end{itemize}
Each fix moved the bottleneck: \emph{when} to think $\rightarrow$ \emph{what} to prescribe $\rightarrow$
\emph{how long} to commit. In the thirty games the chain then loops back to \emph{what}: a prescription
made under partial observability, with game knowledge that is right on one map and wrong on another. We
had read the pilot as pointing next to \emph{how} the agent moves (doomLaya's \emph{explore} did not find
the key on MAP02 within 180\,s); these data do not support that reading.

\section{Discussion and limitations}\label{sec:limits}
\paragraph{One system, two speeds.} We designed the switch as a single system: fast by default, slow
on demand. Our data do not test whether a single model running in two modes (deciding without
reasoning, thinking only when the gate opens) does better than two models. With two models, the offline
evidence supports choosing the fast part for the sensitivity that matters to the deferral and the slow
part for its ability to reason. That sensitivity can be coarse: the confidence of the actor that pays
most tracks which kinds of situation it fails, which is enough when accuracy varies so much between kinds.
\paragraph{What the closed loop says.} Offline accuracy and closed-loop behaviour measure different
things here: the oracle never labels exploring, and in play the useful prescription was mostly to
explore. In the loop, committing to a plan mattered more than its content (with both actors the rule
control covers as much ground; the thinker's plans die less), knowledge given to the reasoner was over-applied where the state could not contradict it,
and thinking occupied up to 70\% of the game, on frames 9--29\,s old. A reasoner in the loop needs
an action vocabulary and an executor that can carry out its plans, plans that last as long as their
goals, and a state that tells it what it needs to know.
\paragraph{Limitations.}
\begin{itemize}
\item Accuracy is agreement with a rule-based oracle that encodes one policy (e.g.\ collect only what is
needed) and never labels exploring, retreating or waiting; a reasonable player can disagree.
\item doomLaya presents the options in a fixed order, and part of several models' accuracy and
sensitivity depends on it (\S\ref{sec:zeroshot}, \S\ref{sec:meta}); the thinkers were evaluated in the
fixed order only.
\item The actor and the deferral rate of the headline numbers are chosen on the test set; the held-out
selection of \S\ref{sec:defer} gives smaller and more honest figures.
\item Offline, the thinkers see the text state only; in play, Qwen3.6 also sees the frame. Without the
frame it reasons longer and covers less ground, but we did not measure offline what the frame adds.
\item The closed loop is small: 33 games on one map, three seeds per variant, no exit, so we report
behaviour, not success rates; differences in deaths may be noise. The single-seed pilot on two maps
changed several settings per step and was run with an abliterated (refusal-removed) variant of
Qwen3.6-35B-A3B; everything else uses official weights.
\item We did not measure systematically whether the thoughts are faithful to the game state; we only
checked individual cases (one apparent mismatch was an item id reused by the game).
\item Latency of reasoning: with basics and commitment a median of 9--29\,s per thought and up to 32\,s,
during which the actor plays on; between 3 of 43 (bare thinker) and 9 of 16 (no image) thoughts exceeded
the 4{,}000-token budget and were answered without reasoning. Offline, Qwen3.6 left 13 of 600 questions
unanswered within the budget and Gemma~4 199; answering those without reasoning, as the switch does,
raises Qwen3.6 from 0.69 to 0.71 and Gemma~4 from 0.50 to 0.61, and the gains of
Table~\ref{tab:defer} barely change (0.75 and 0.69 at 30\% for the Laya base).
\end{itemize}

\section{Conclusion}
A fast decision model whose confidence tracks which situations it fails, a reasoner asked only then, and
plans that last as long as their goals: this is the System Switch. Offline, deferring the least
confident decisions to a reasoning model gains over deferring at random in proportion to the actor's
sensitivity, without any training; the most accurate zero-shot actors are not the ones to use. In real
time, the games turn a vague ``the agent fails'' into a precise list of what fails, down to a door that
the reasoner believes locked because we told it that doors can be. They also set the bar for what
follows: a fixed rule behind the same gate covers as much ground, so in this loop the thoughts have so
far bought fewer deaths, not progress towards the exit.

\section*{Acknowledgements and use of AI}
The experiments were designed by the author. Code, data analysis and a first draft of the text were
produced with the help of an AI assistant (Claude, Anthropic), under the author's direction and review;
all numbers come from the released data.

\section*{Reproducibility}
Code, prompts and data are available at
\begin{center}\url{https://github.com/dexmac221/system-switch}\end{center}
\begin{sloppypar}
The repository contains the switch (\texttt{switch.py}), the bridge to \texttt{/v1/systemone}
(\texttt{serve\_systemone.py}),
the test set builder (\texttt{build\_testset.py}), the evaluation scripts, and the scripts that produce
every table and figure from the per-question records (\texttt{paper\_tables.py},
\texttt{review\_analyses.py}, \texttt{loop\_metrics.py}), the patches to doomLaya, and in
\texttt{data/} the test set (in both option orders), the per-question records of every actor and
thinker, and the switch logs, summaries and per-tic positions of every game. The game videos are
available from the author.
\end{sloppypar}

\bibliographystyle{plainnat}
\bibliography{refs}
\end{document}